\documentclass{article}

\usepackage{multirow}
\usepackage[preprint,nonatbib]{neurips_2023}
\usepackage[hyphens]{url}  

\usepackage[utf8]{inputenc} 
\usepackage[T1]{fontenc}    
\usepackage{hyperref}       
\usepackage{url}            
\usepackage{booktabs}       
\usepackage{amsfonts}       
\usepackage{nicefrac}       
\usepackage{xcolor}   
\usepackage{graphicx}
\usepackage{wrapfig}
\usepackage{makecell}
\usepackage{comment}
\usepackage{amsmath}
\usepackage{lmodern}
\usepackage{booktabs} 

\usepackage[most]{tcolorbox}
\tcbset{colback=gray!5!white, colframe=gray!75!black, boxrule=0.5pt}

\usepackage{longtable}
\usepackage{array}
\usepackage{tabularx}

\usepackage{graphicx}
\usepackage{amsthm}
\usepackage{mathtools}
\usepackage{amssymb}
\usepackage[makeroom]{cancel}

\usepackage[config,font=small]{caption,subfig}
\usepackage{parskip}
\usepackage{setspace}
\numberwithin{equation}{section}
\numberwithin{figure}{section}
\usepackage{enumerate}
\usepackage{float}

\usepackage[colorinlistoftodos,textsize=tiny]{todonotes}

\newcommand\be{\begin{equation}}
\newcommand\bea{\begin{eqnarray}}
\newcommand\ee{\end{equation}}
\newcommand\eea{\end{eqnarray}}

\title{FeynTune: Large Language Models for High-Energy Theory}

\author{%
\textbf{Paul Richmond}$^{\spadesuit}$, \textbf{Constantinos Papageorgakis}$^{\spadesuit}$\thanks{Corresponding author.}\;, \textbf{Vasilis Niarchos}$^{\diamondsuit}$\;, \\
\textbf{Borun Chowdhury}$^{\clubsuit}$,\quad \textbf{Prarit Agarwal}$^{\heartsuit}$ \quad 
 \\ 
\quad 
  \\ \\
$^\spadesuit$ Centre for Theoretical Physics, Department of Physics and Astronomy, \\
Queen Mary University of London, London E1 4NS, UK \\  
$^\diamondsuit$ ITCP \& CCTP, Department of Physics,\\ 
University of Crete, 71003 Heraklion, Greece \\
$^\clubsuit$ Meta, 10 Brock Street, Regents Place, London NW1 3FG, UK\\ 
$^\heartsuit$ Amazon, 1 Principal Place, Worship St, London EC2A 2FA, UK  \\ 
\\
\textbf{Email:} paul\_richmond@ymail.com, 
c.papageorgakis@qmul.ac.uk, \\
niarchos@physics.uoc.gr,
borundev@meta.com,
prarit@amazon.co.uk
}

\begin{document}

\vspace*{-2cm}
\begin{flushright}
\small
CCTP-2025-7 \hfill QMUL-PH-25-13 \\
ITCP-IPP-2025/7
\end{flushright}
\vspace{1cm}

\maketitle
\setcounter{footnote}{0}

\begin{abstract}\small
We present specialized Large Language Models for theoretical High-Energy Physics, obtained as 20 fine-tuned variants of the 8-billion parameter Llama-3.1 model. Each variant was trained on arXiv abstracts (through August 2024) from different combinations of hep-th, hep-ph and gr-qc. For a comparative study, we also trained models on datasets that contained abstracts from disparate fields such as the q-bio and cs categories. All models were fine-tuned using two distinct Low-Rank Adaptation fine-tuning approaches and varying dataset sizes, and outperformed the base model on hep-th abstract completion tasks. We compare performance against leading commercial LLMs (ChatGPT, Claude, Gemini, DeepSeek) and derive insights for further developing specialized language models for High-Energy Theoretical Physics.

\end{abstract}

\section{Introduction}

Foundation models \cite{bommasani2022opportunitiesrisksfoundationmodels}, like the flagship Large Language Models (LLMs) GPT-4 of Open-AI \cite{openai2024gpt4technicalreport}, and the Llama models \cite{touvron2023llamaopenefficientfoundation, touvron2023llama2openfoundation, dubey2024llama3herdmodels} of Meta, exhibit remarkable performance on a wide range of tasks (from logical reasoning to creative writing), gaining a lot of traction in recent years over diverse domains, with both commercial and scientific applications.\footnote{For a more complete list of popular LLMs, the reader can consult, for example, the blog post \href{https://explodingtopics.com/blog/list-of-llms}{https://explodingtopics.com/blog/list-of-llms} and the GitHub page \href{https://github.com/Hannibal046/Awesome-LLM}{https://github.com/Hannibal046/Awesome-LLM}.} For the scientific community, in particular, the ability of LLMs to accelerate the scientific process and discovery is a topic of obvious interest. For example, a comprehensive survey of over 260 scientific LLMs in Mathematics and the Natural Sciences \cite{zhang2024comprehensivesurveyscientificlarge} summarizes applications that range from the generation of novel scientific ideas to problem-solving in Mathematics, Physics, Chemistry and Materials Science, Computer Science, Biology and Medicine, and Environmental Sciences; see also \cite{bubeck2025earlysciaccelgpt5} for very recent results in this direction, where some new mathematical results were obtained with the use of ChatGPT-5. 

Different scientific fields have embraced this approach to different degrees. In particular, we are aware of only a handful of Physics-trained LLMs. These include astroBERT \cite{grezes2021buildingastrobertlanguagemodel}, AstroLlama/AstroMLab \cite{Nguyen:2023nhp, UniverseTBD:2024pmh, ting2024astromlab1winsastronomy}, AstroLlama-chat \cite{perkowski2024astrollamachatscalingastrollamaconversational}, CosmoSAGE \cite{dehaan2024cosmosagenaturallanguageassistantcosmologists} and PhysBERT \cite{hellert2024physberttextembeddingmodel}. The latter is a physics-specific model for sentence embedding. Compared to other domains, this is a relatively short list, primarily focused on Astronomy and Cosmology. One of the aims of this paper is to augment this list by taking the first steps towards the construction of LLMs that will facilitate research in Theoretical Physics, with primary focus on Theoretical High-Energy Physics and adjacent fields like Classical and Quantum Gravity, and High-Energy Physics Phenomenology. 

{\it Why Theoretical Physics?} First, we are motivated by our own research interests and scientific background, ultimately wanting to create a domain-specific LLM that will facilitate research in High-Energy Theory. This can occur by providing reliable guidance to the literature and useful assistance to problem-solving that goes beyond the abilities of standard computational packages (like Wolfram's Mathematica), as well as the generation of new ideas. Eventually, one would like to move towards a conversational AI assistant that combines interdisciplinary intuition not only from the existing high-energy theory (hep-th) literature, but also from other neighboring fields, {\it e.g.,} High-Energy Lattice (hep-lat), General Relativity (gr-qc), High-Energy Phenomenology (hep-ph), High-Energy Experiment (hep-ex), Condensed Matter Theory (cond-mat) and Mathematical Physics (math-ph).

Second, we believe that Theoretical Physics presents rich opportunities for future developments in AI. On the one hand, it lies close to pure Mathematics: it is heavily based on mathematical language and reasoning; it values mathematical rigor and consistency. On the other hand, Theoretical Physics, being based on natural phenomena, also involves physics-based hypotheses, reasoning and consistency checks which do not directly follow from mathematics alone (e.g. one cannot arrive at the principle of wave-particle duality or Schrodinger's wave-equation via mathematics alone). While a significant push is being made to equip LLMs with mathematical reasoning capabilities, we believe a similar effort in teaching physics-based reasoning and logic to LLMs will help them build a better internal model of the world.

With this motivation, in this paper we undertake the first steps towards the construction of LLMs  fine-tuned on the existing literature of theoretical High-Energy Physics. Our goals are modest and meant to serve as a proof of concept that will spur further future developments. The models were trained only on arXiv paper abstracts and their  performance tested on abstract completion. We also chose a base model with a relatively small number of parameters: the 8-billion (8B) version of one of the more recent Llama models \cite{dubey2024llama3herdmodels}. In order to strike a balance between domain expertise and the ability to cross-fertilize ideas across adjacent domains of Science, we fine-tuned a number of different variants that involved: a pure High-Energy Theory (hep-th) model and combinations of hep-th with High-Energy Phenomenology (hep-ph) and Classical and Quantum Gravity (gr-qc) models with different training sample sizes. We also fine-tuned the base model on combinations of datasets involving samples lying farther afield, such as Computer Science (cs) and Quantitative Biology (q-bio). We compare the performance of these models with each other and against the base Llama model. For all variants we trained two versions: one where Low-Rank Adaptation (LoRA) adapters were only applied to the Query-Key-Value  matrices (LoRA-QKV), and a second where LoRA adapters were applied to all projection matrices (LoRA-all). 

A summary of our key findings follows:

\begin{enumerate}
    \item All fine-tuned models trained on hep-th abstracts produce better completions compared to the base model.
    \item Given that hep-th is a field with a relatively small repository of papers, performance can be improved by augmenting training datasets  with papers from other domains. Interestingly, models trained on more diverse datasets are prone to producing more creative completions.     
    \item Solely training on adjacent, non-hep-th, field abstracts results in worse performance on the hep-th test dataset compared to our other fine-tuned models, as expected, but still better than the base model in terms of human evaluation.
    \item The LoRA-all collection of  models exhibited an unusual step-function profile for the training loss. Such curves have been observed before \cite{howard2023learningjumps} and we found that they do not impact model performance.
    \item Comparisons of completion samples between our fine-tuned variants and the free versions of some popular commercial models reveal that our models can produce comparable completions with frequent use of technical, expert language, but limited factual accuracy.

\end{enumerate}

\section{Datasets and Training}

Our starting point for fine-tuning was Meta's Llama 3.1 8B foundation model \cite{dubey2024llama3herdmodels} due to its small size, open weight nature and wide integration into Python libraries.\footnote{The model can be downloaded from Hugging Face at \href{https://huggingface.co/meta-llama/Llama-3.1-8B}{https://huggingface.co/meta-llama/Llama-3.1-8B}.} We note that Llama models already use arXiv metadata as part of their training datasets, including abstracts \cite{touvron2023llamaopenefficientfoundation,lewkowycz2022solvingquantitativereasoningproblems}.

We fine-tuned this base model on a collection of 10 abstract datasets, which we denote s1-s10 for brevity and summarize in Table~\ref{tab:dataset-summary}. These were curated from publicly available arXiv metadata,\footnote{The metadata can be accessed at: \href{https://www.kaggle.com/datasets/Cornell-University/arxiv}{https://www.kaggle.com/datasets/Cornell-University/arxiv}.} including  all preprint submissions up to late August 2024.

\begin{table}[t]
\centering
\begin{tabular}{clr}
    \toprule
    \textbf{Abbrev} & \textbf{Datasets} & \textbf{Abstracts \#} \\
    \midrule
    s1   & hep-th                   & 105,384 \\
    s2   & hep-ph, gr-qc           & 195,909 \\
    s3   & hep-th, hep-ph, gr-qc   & 301,293 \\
    s4   & hep-th, hep-ph, gr-qc   & 105,384 \\
    s5   & hep-th, gr-qc           & 105,384 \\
    s6   & hep-th, hep-ph          & 105,384 \\
    s7   & hep-th, gr-qc           & 168,897 \\
    s8   & hep-th, hep-ph          & 237,780 \\
    s9   & hep-ph, gr-qc           & 105,384 \\
    s10  & hep-th, q-bio, cs       & 301,293 \\
    \bottomrule
\end{tabular}
\vspace{0.5em}
\caption{Summary of curated dataset abbreviations and abstract counts. These are publicly available at \href{https://huggingface.co/LLMsForHepth}{https://huggingface.co/LLMsForHepth}.}
\label{tab:dataset-summary}
\end{table}

Our list included datasets matching the size of the full hep-th category (s1) with different percentages of non-hep-th content (s4, s5, s6, s9), as well as concatenations of all available abstracts in a given category (s2, s3, s7, s8) to test the correlation between model performance and dataset size. The last dataset (s10) was created by combining all hep-th and q-bio abstracts, as well as enough cs abstracts to match the size of our largest dataset, s3.  The data were randomly shuffled and divided into train/validation/test sub-datasets using a 70\%-15\%-15\% split. We detail the composition of samples per arXiv category for each dataset in Appendix~\ref{curation}.  

The fine-tuning process was carried out using the Hugging Face ecosystem. In order to improve resource efficiency, the Llama model weights were quantized to 4-bit precision with the bitsandbytes package \cite{dettmers2023case,dettmers2023qlora} and we employed LoRA \cite{hu2022lora} to reduce the number of trainable parameters. LoRA works by freezing the pre-trained model weights and inserting trainable low-rank matrices into each layer, allowing efficient adaptation with a small fraction of the original parameters—--this is analogous to treating the fine-tuning process as a perturbation on the base model's weight matrices. As mentioned in the Introduction, we experimented with two versions of LoRA: one applied to the QKV matrices only (LoRA-QKV), and one applied to all projection matrices (LoRA-all).  A detailed hyperparameter list is given in Table~\ref{tab:training-config}, with the complete settings presented in Appendix \ref{fullhypers}.

\begin{table}[t]\small
\centering
\begin{minipage}{0.48\linewidth}
\centering
\begin{tabular}{@{}ll@{}}
\toprule
\textbf{Component} & \textbf{Setting} \\
\midrule
Optimizer & AdamW \\
AdamW $\beta_1$, $\beta_2$ & 0.9, 0.95 \\
Weight decay & 0.1 \\
Learning rate & $3 \times 10^{-4}$ (max), $3 \times 10^{-6}$ (min) \\
Schedule & Linear warmup (10\%), \\
& then cosine decay \\
& \\
\bottomrule
\end{tabular}
\vskip 0.25cm
\caption*{(a) Optimizer and learning rate schedule}
\end{minipage}
\hfill
\begin{minipage}{0.48\linewidth}
\centering
\begin{tabular}{@{}ll@{}}
\toprule
\textbf{Component} & \textbf{Setting} \\
\midrule
LoRA rank ($r$) & 8 \\
LoRA scaling factor ($\alpha$) & 32 \\
Dropout & 0.05 \\
LoRA modules & \texttt{up\_proj}, \texttt{down\_proj},\\
             & \texttt{k\_proj}, \texttt{q\_proj}, \\
             & \texttt{v\_proj}, \texttt{o\_proj} \\
             &\texttt{gate\_proj}, \\
\bottomrule
\end{tabular}
\vskip 0.25cm
\caption*{(b) LoRA configuration}
\end{minipage}
\vskip 0.25cm
\caption{Training configuration: optimizer settings and model-specific hyperparameters.}
\label{tab:training-config}
\end{table}

All models were trained over four epochs using three NVIDIA A100 40Gb GPUs with a (per device) batch size of 16. This hardware allowed us to take advantage of bfloat16 16-bit mixed precision and Flash Attention 2 \cite{dao2023flashattention2} for the attention mechanism. The loss function was taken to be the standard cross-entropy loss. During the four epochs of training we ran ten evaluations on the entire train and validation datasets, including at the start and end of training. As part of these evaluations, we also recorded the  perplexity per batch for the dataset.

At this point we would like to pause and discuss a potential ambiguity in the definition of perplexity. While the definition of perplexity for one sequence is the exponential of the cross entropy loss, it is not a priori clear how the perplexity of a dataset, consisting of multiple sequences, is defined. We have seen instances that seem to suggest that one should take the geometric mean~\cite{huyen2024ai}. However, we are of the opinion that the more relevant metric is the arithmetic mean; see Appendix~\ref{app:cel_perplexity} for a related discussion. In this paper when we discuss perplexity over a dataset, we mean the arithmetic mean of the perplexities of the individual sequences.

\begin{figure}[t]
    \centering
    \begin{minipage}{0.48\textwidth}
        \centering
        \includegraphics[width=1.0\linewidth]{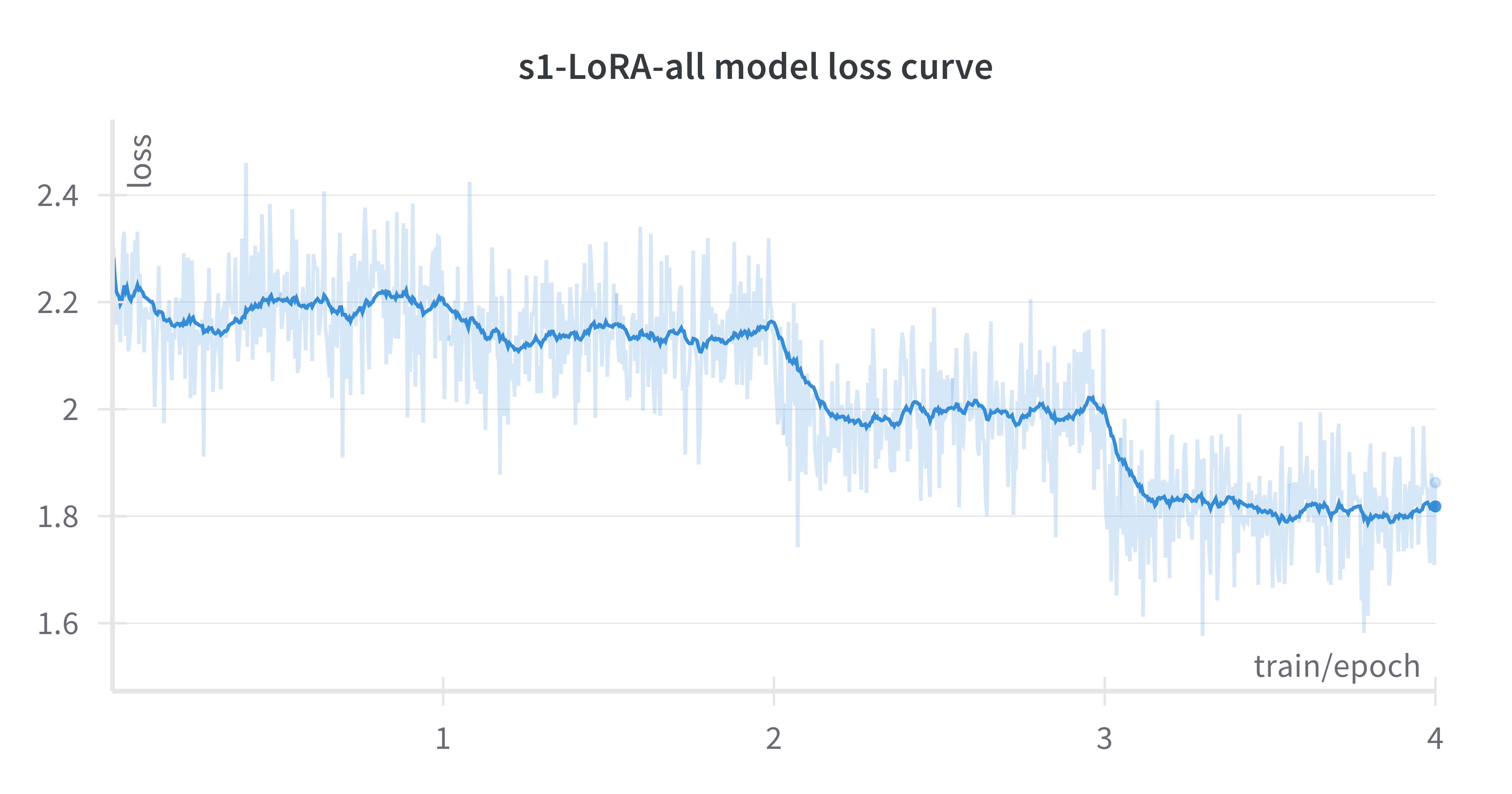}
    \end{minipage}
    \hfill
    \begin{minipage}{0.48\textwidth}
        \centering
        \includegraphics[width=1.0\linewidth]{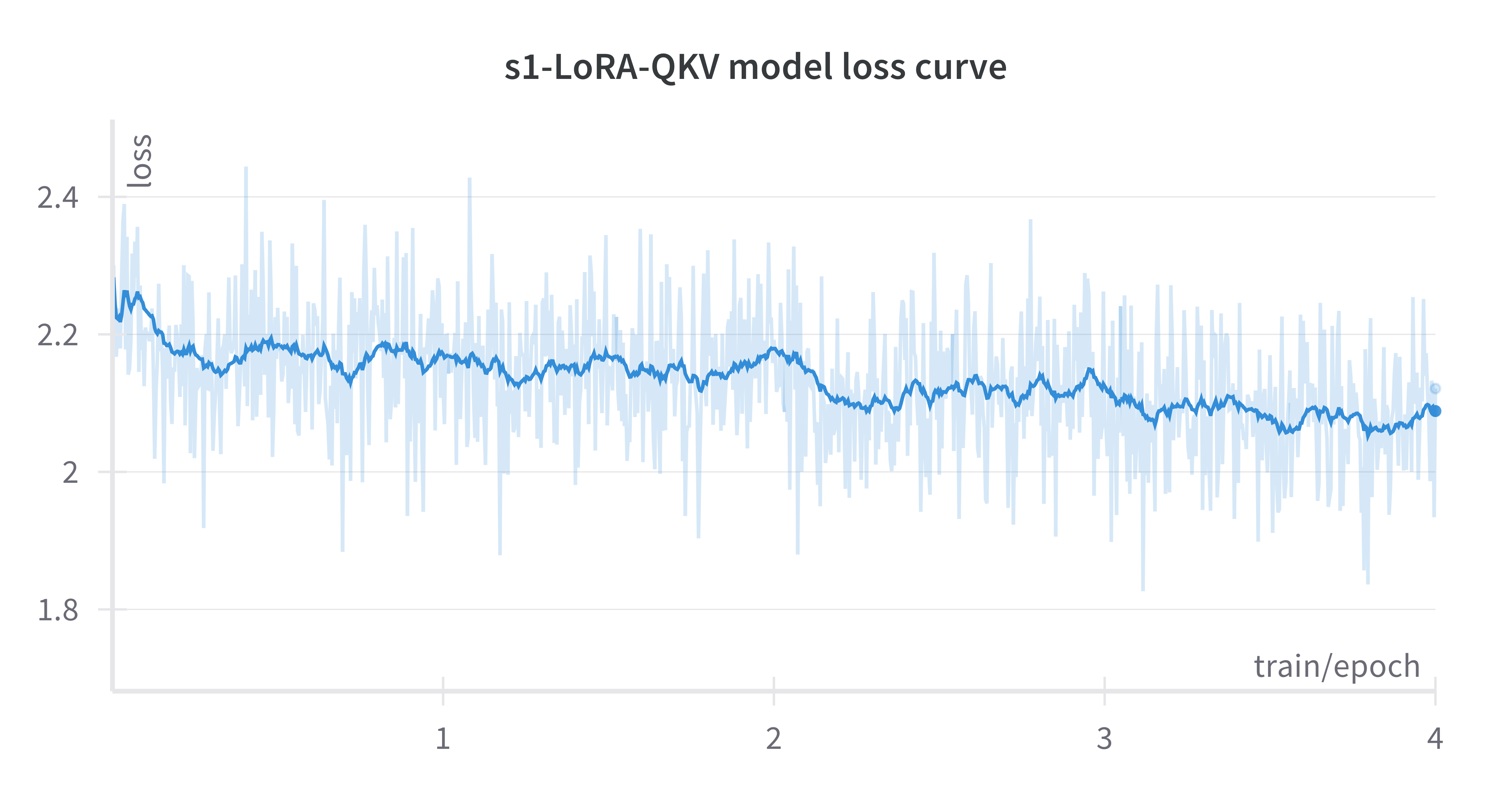}
    \end{minipage}
    \caption{Learning curves comparing the s1-LoRA-all model (left) showcasing a step-wise decrease in loss with the s1-LoRA-qkv model (right) showing a standard loss function behavior.}
    \label{fig:TH_learning_curves}
\end{figure}

\begin{figure}[t]
    \centering
    \begin{minipage}{0.48\textwidth}
        \centering
        \includegraphics[width=1.0\linewidth]{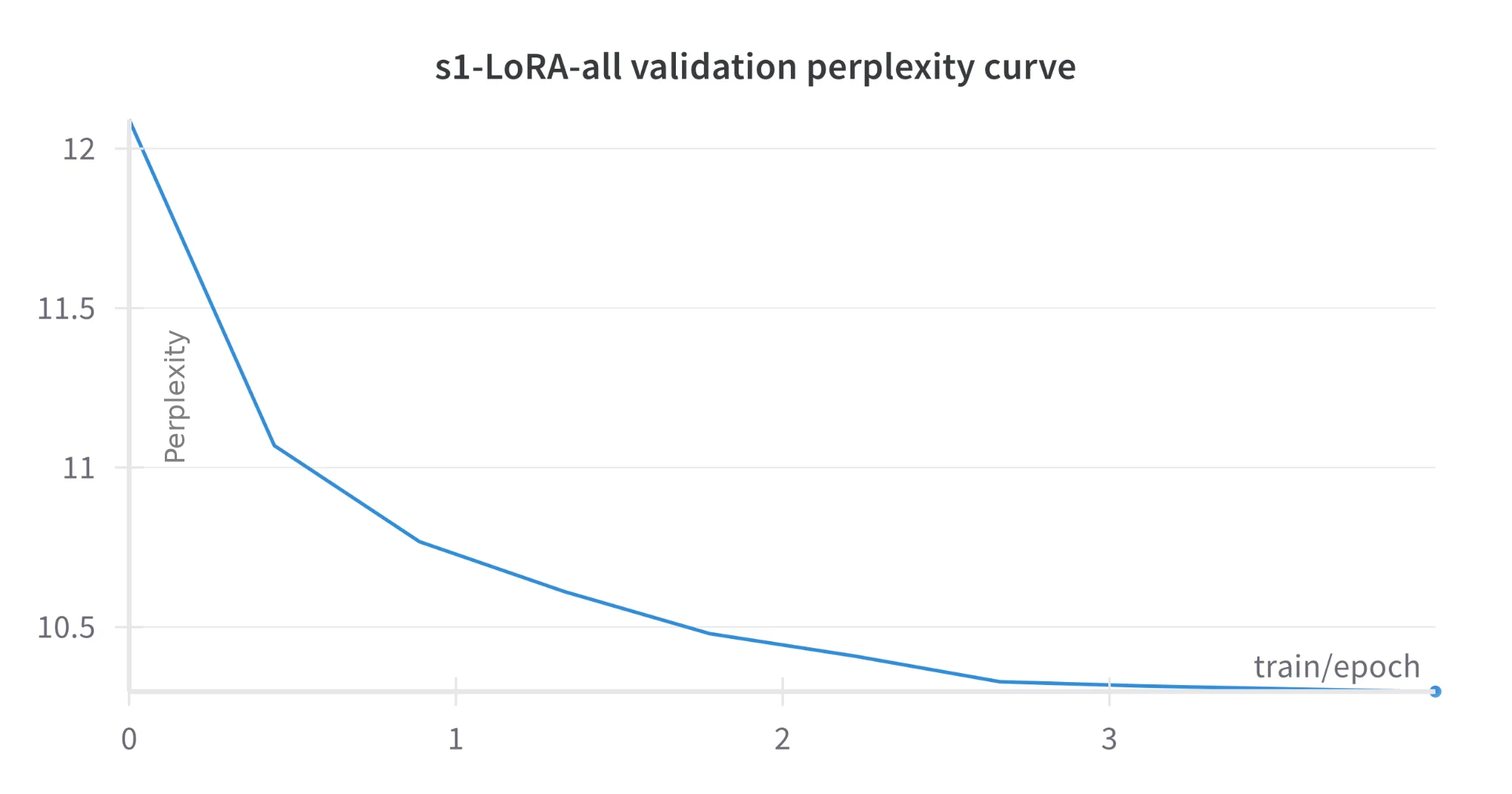}
    \end{minipage}
    \hfill
    \begin{minipage}{0.48\textwidth}
        \centering
        \includegraphics[width=1.0\linewidth]{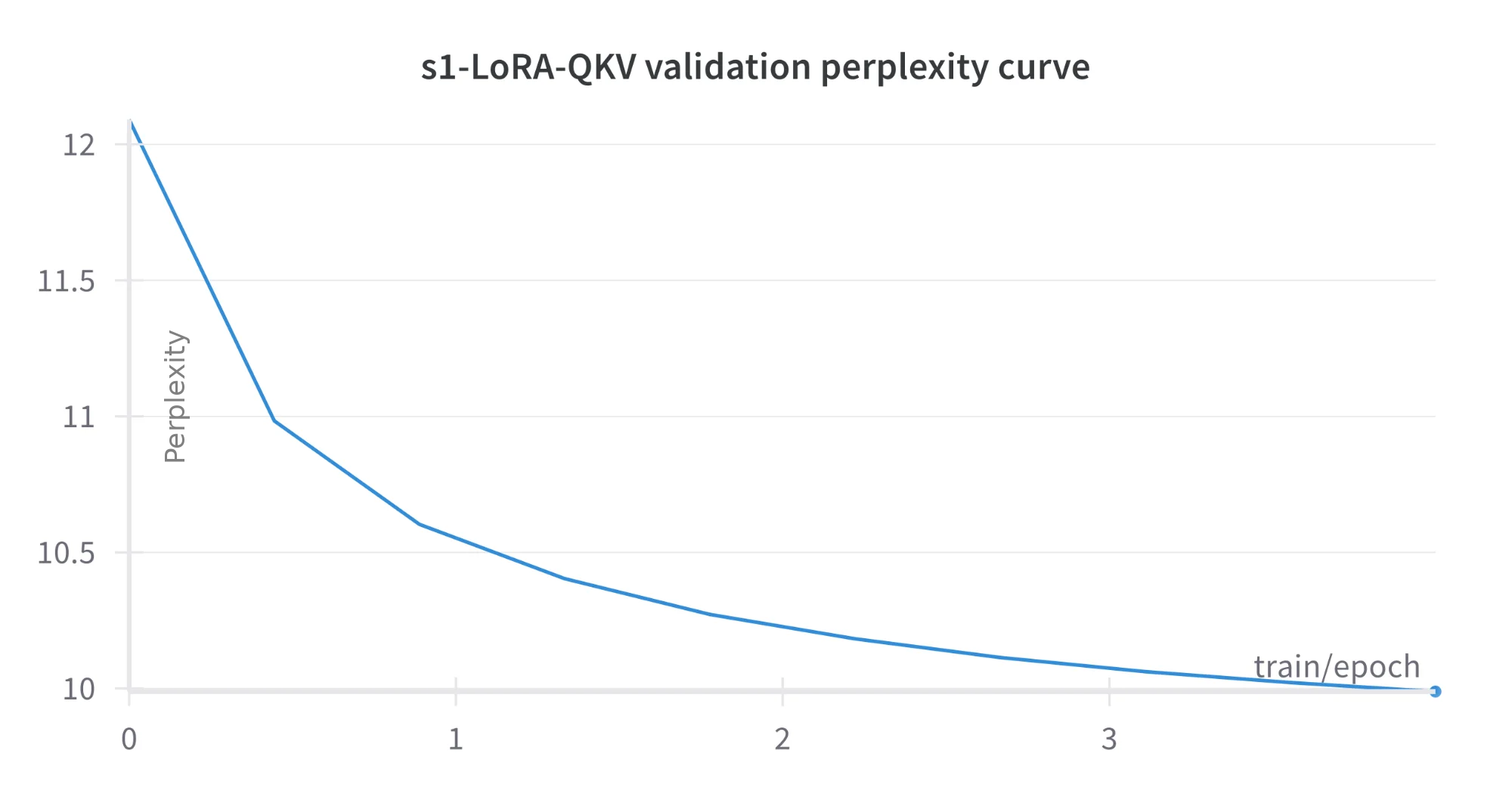}
    \end{minipage}
    \caption{Smoothed mean batch perplexity evaluation curves on the s1 validation dataset, comparing the s1-LoRA-all model (left) and the s1-LoRA-QKV model (right).}
    \label{fig:TH_perplexity_curves}
\end{figure}

While monitoring training, we noticed that the learning curves for the LoRA-all models exhibited an intriguing feature: the loss was rather constant over each epoch, with a step-function drop between different epochs; this was typical for all models s1-s10. This behavior has been observed before---see e.g.\ \cite{howard2023learningjumps}---and we do not find it to be pathological; see Figures~\ref{fig:TH_learning_curves} and \ref{fig:TH_perplexity_curves}.\footnote{These models achieve lower training loss after 4 epochs compared to the LoRA-QKV models that exhibit a more standard training curve. We note that for the evaluation curves the LoRA-QKV models were on average slightly better performing.} We will also report momentarily that there were no standout differences between LoRA-all and LoRA-QKV models during the human evaluation phase. 

The code used to curate the datasets, fine-tune and evaluate the models can be found at \href{https://github.com/Paul-Richmond/FeynTune}{https://github.com/Paul-Richmond/FeynTune}. The fine-tuned models are available as LoRA adapters at \href{https://huggingface.co/LLMsForHepth/models}{https://huggingface.co/LLMsForHepth/models}.

\section{Evaluation Metrics}

To evaluate the performance of the fine-tuned models, we processed each abstract in the target hep-th test dataset (denoted s1) by splitting it into a prompt and ground truth as follows. Each abstract was first divided into sentences, with the total number of sentences in an abstract denoted by 
$N$.\footnote{If an abstract contained only a single sentence, it was instead split into $N$ words.} The prompt was constructed by taking the first $\lceil \frac{N}{2} \rceil$ sentences, while the remaining sentences formed the ground truth. The prompts were then run through the models fine-tuned on the datasets of Table~\ref{tab:dataset-summary}---which we will refer from here on by the same abbreviation s1-s10---to obtain completions with temperature 0.8 and a maximum of 1024 tokens. We point out that during training we quantized the base LLaMA model weights to 4-bit precision and fine-tuned 16-bit LoRA adapters on top. However, for evaluation we merged the trained LoRA weights back into the base model and loaded the entire merged model in 16-bit precision. The full settings used for inference can be found in Appendix \ref{fullhypers}.

\subsection{Perplexity}

For each of our fine-tuned models, including the LoRA-QKV and LoRA-all variants, we evaluated the mean batch perplexity on the hep-th (s1) test dataset. In order to get a comparative sense between models, we also computed the mean and standard deviations of these perplexities using the bootstrap technique. In this technique the sampling distribution of a statistic is estimated by resampling with replacement from the original dataset. Statistics are then extracted by computing the arithmetic mean and standard deviation of the bootstrap distribution. The results are displayed in Table~\ref{tab:perplexity-stats}, with a visual presentation given in Figure~\ref{fig:perp_boot_comparison}. For reference, Transformer-XL achieved 18.9 perplexity on WikiText-103~\cite{dai2019transformerxl}. Although our fine-tuned models performed better on hep-th abstracts, direct comparison is complicated by differences in tokenizer vocabulary size, aggregation conventions (arithmetic vs. geometric mean), and domain characteristics--—scientific abstracts are more formulaic than Wikipedia text. Therefore, a more meaningful metric is relative improvement.

Unsurprisingly, all LoRA-all models fine-tuned on datasets including hep-th abstracts (s1, s3-s8, s10) exhibited lower perplexity compared to the base model. On the contrary, models that did not include hep-th abstracts (s2, s9) had similar/higher perplexity. We also noted that the overall size of the dataset matters, with the lowest perplexity variants being those that included all hep-th samples augmented with other datasets (hep-ph, gr-qc, cs, q-bio). 

Turning to the LoRA-QKV models, the perplexity was overall lower than for LoRA-all, with even the s2 and s9 models scoring lower than the base model. The remaining LoRA-QKV models were comparable, with no statistically significant differences. Comparing LoRA-all and LoRA-QKV trained variants of the same dataset, we note that the QKV version almost always has equal or lower perplexity. This fact, along with the observations relating to the structure of the learning curves, initially biased us towards the QKV variants performing better. However, we will argue that human evaluation did not support this expectation.

\begin{table}[t]
\centering
    \renewcommand{\arraystretch}{1.5}
\begin{tabular}{@{}lrr@{}}
\toprule
\textbf{Model} & \textbf{LoRA-all} & \textbf{LoRA-QKV} \\
\midrule
Llama & $11.20^{+0.16}_{-0.13}$ & $11.20^{+0.16}_{-0.13}$ \\
    \midrule
    s1 & $10.32^{+0.11}_{-0.09}$ & $10.04^{+0.13}_{-0.10}$ \\
    s2 & $11.38^{+0.11}_{-0.10}$ & $10.62^{+0.15}_{-0.12}$ \\
    s3 & $9.83^{+0.11}_{-0.09}$ & $10.02^{+0.14}_{-0.11}$ \\
    s4 & $10.41^{+0.11}_{-0.10}$ & $10.08^{+0.12}_{-0.10}$ \\
    s5 & $10.35^{+0.11}_{-0.09}$ & $10.07^{+0.12}_{-0.10}$ \\
    s6 & $10.34^{+0.09}_{-0.08}$ & $10.05^{+0.13}_{-0.11}$ \\
    s7 & $10.00^{+0.10}_{-0.09}$ & $10.04^{+0.14}_{-0.12}$ \\
    s8 & $9.90^{+0.08}_{-0.08}$ & $10.04^{+0.12}_{-0.10}$ \\
    s9 & $12.16^{+0.28}_{-0.19}$ & $10.64^{+0.14}_{-0.11}$ \\
    s10 & $9.93^{+0.11}_{-0.09}$ & $10.00^{+0.15}_{-0.11}$ \\
\bottomrule
\end{tabular}
    \renewcommand{\arraystretch}{1.0}
\vskip 0.5cm
\caption{Comparison of perplexity means across base and fine-tuned model configurations. Lower perplexity indicates better model performance. The perplexity of the base Llama model without any fine-tuning (LoRA-all or LoRA-QKV) has been included to make the comparison to the baseline easier. Uncertainties shown as $^{+\text{upper}}_{-\text{lower}}$ represent 95\% confidence intervals from bootstrap resampling.}
\label{tab:perplexity-stats}
\end{table}
\subsection{Semantic Similarity}
\label{cosine}

As an additional performance measure, we evaluated semantic similarity by computing embedding vectors for all abstract completions on the hep-th test dataset. We used the SemScore model ({\tt sentence-transformers/all-mpnet-base-v2}), \cite{2024arXiv240117072A}, which produces 768-dimensional embeddings, to encode both the model predictions and ground truth abstracts. The cosine similarity between these embedding vectors provides a quantitative measure of semantic alignment between fine-tuned model outputs and ground truth abstracts.

On average we observed that the base model scored $0.88 \pm 0.08$ while all the fine-tuned models (LoRA-QKV and LoRa-all) $0.90 \pm 0.07$. Therefore, from the perspective of SemScore all models (base and fine-tuned) exhibit comparable performance in semantic similarity compared to the ground truth (with the fine-tuned models perhaps having a marginal edge). We note that the minimum values of the cosine similarities exhibited more pronounced variation: for the base model the minimum was 0.007, for the LoRA-all models the minimum was above 0.1 (with the exception of the s2 model, that was not trained on hep-th, at 0.06), and for the LoRA-QKV models the minimum had a much higher variation ranging from $-0.015$ to 0.15. For example, the s10 LoRA-QKV model (trained on hep-th, q-bio, cs) had minimum cosine similarity at $-0.001$. 

Under closer inspection of randomly chosen abstracts, we could not identify any correlations between semantic similarity (as perceived by the embedding of SemScore) with the overall quality of the completion. The above cosine similarity numbers simply confirm that the fine-tuned models have inherited successfully the ability of the base model to produce semantically relevant completions.

\begin{table}[t]
\centering
\begin{tabular}{lcccc}
\toprule
Category & Mean & SD & Median & $N$ \\
\midrule
Base & 3.75 & 1.52 & 4.0 & 36 \\
LoRA-all & 5.43 & 1.74 & 6.0 & 360 \\
LoRA-QKV & 4.92 & 1.70 & 5.0 & 360 \\
Commercial & 7.50 & 1.99 & 7.5 & 144 \\
\bottomrule
\end{tabular}
\vskip 0.1cm
\caption{Human evaluation scores by model category.}\label{tab:summary_stats}
\end{table}

\subsection{Human evaluation} 
\label{human}

To assess the quality of model-generated abstract completions, we conducted a human evaluation study with three domain experts, all experienced researchers in high-energy theoretical physics (hep-th). Each evaluator independently rated completions on a scale of 1--10 according to the following criteria:

\begin{itemize}
    \item \textbf{1--3 (Poor):} Output is nonsensical, contains obvious errors, or is entirely generic with no connection to the specific physics context of the prompt.
    \item \textbf{4--6 (Acceptable):} Output demonstrates some domain awareness but may contain inaccuracies, lack specificity, or fail to coherently continue the abstract's argument.
    \item \textbf{7--9 (Good):} Output is coherent, scientifically plausible, and appropriately continues the abstract with relevant physics content.
    \item \textbf{10 (Excellent):} Output is indistinguishable from a genuine hep-th abstract continuation, demonstrating deep domain understanding.
\end{itemize}

The evaluation corpus comprised 25 abstracts spanning diverse hep-th topics (string theory, conformal field theory, black holes, gauge/gravity duality), with completions generated by 25 model variants: the base Llama-3.1-8B model, 10 LoRA-all fine-tuned variants, 10 LoRA-QKV fine-tuned variants, and 4 commercial models (ChatGPT-4, Claude, Gemini, DeepSeek). This yielded 900 total human ratings, with 5 abstracts evaluated by all three raters, 1 by two and 19 individually.\footnote{The human evaluation scores are also available as a CSV file in the GitHub repository \href{https://github.com/Paul-Richmond/FeynTune}{https://github.com/Paul-Richmond/FeynTune}.} 

\paragraph{Statistical Analysis:}
Table~\ref{tab:summary_stats} presents summary statistics by model category for all 900 ratings. Fine-tuned models significantly outperformed the base model (Mann-Whitney $U = 6{,}947$, $p < 0.001$, rank-biserial $r = 0.46$), while commercial models outperformed fine-tuned models ($U = 19{,}764$, $p < 0.001$, $r = 0.62$). The two LoRA variants showed a small but statistically significant difference ($U = 76{,}754$, $p < 0.001$, $r = 0.18$), with LoRA-all marginally outperforming LoRA-QKV.

\paragraph{Inter-Rater Reliability:}
To assess evaluation consistency, we analyzed the 125 (abstract, model) pairs rated by all three evaluators. Pairwise Spearman correlations ranged from $\rho = 0.47$ to $\rho = 0.59$ (all $p < 0.001$), indicating moderate agreement on relative rankings. The intraclass correlation coefficient ICC(2,1) $= 0.49$ [95\% CI: 0.30--0.64] reflects fair single-rater reliability, while ICC(2,k) $= 0.74$ [95\% CI: 0.56--0.84] indicates acceptable reliability when averaging across raters~\cite{koo2016guideline}. Agreement rates were 21\% exact, 54\% within $\pm 1$ point, and 77\% within $\pm 2$ points---consistent with the inherent subjectivity of evaluating scientific text quality.

In addition to the above metrics, for which we give a qualitative description in Appendix \ref{statmetrics}, we also report on some notable qualitative observations:

\begin{figure}[t]
    \centering
    \begin{minipage}{0.48\textwidth}
        \centering
        \includegraphics[width=1.0\linewidth]{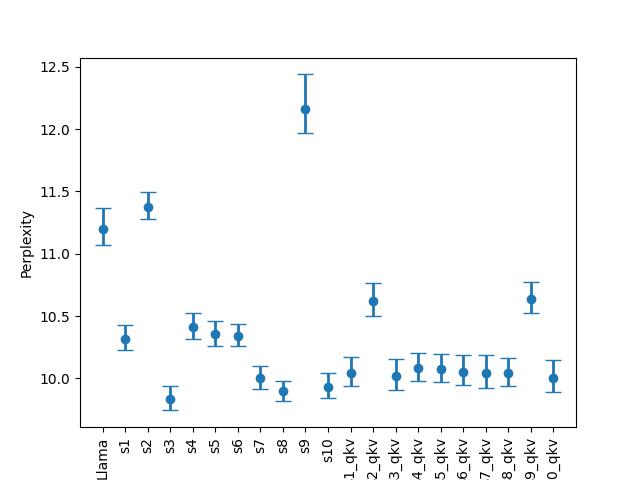}
    \end{minipage}
    \hfill
    \begin{minipage}{0.48\textwidth}
        \centering
        \includegraphics[width=1.0\linewidth]{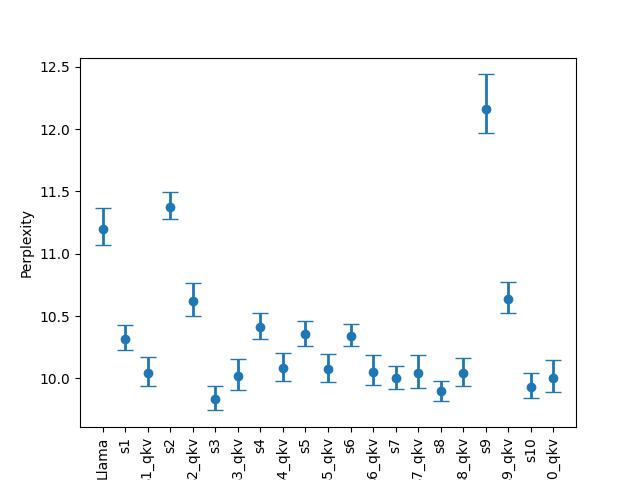}
    \end{minipage}
    \caption{Bootstrapping the perplexities of trained models on the s1 test dataset: standard grouping (left) and paired comparison of  LoRA-all vs LoRA-QKV models (right). Note that the perplexity for the base model is without fine-tuning (LoRA-all or LoRA-QKV); we have included these values  to facilitate  the comparison to the baseline.}
    \label{fig:perp_boot_comparison}
\end{figure}

\begin{figure}[t]
        \centering
        \includegraphics[width=1.0\linewidth]{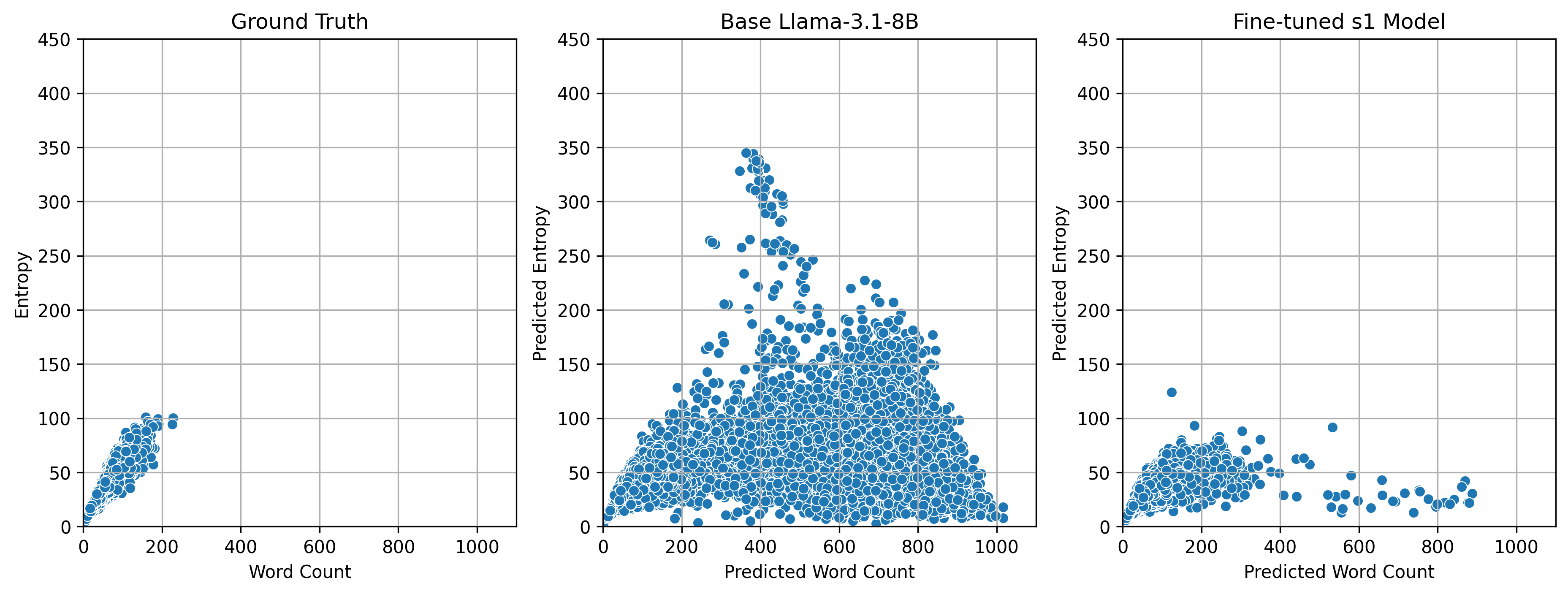}
    \caption{Exponentiated Shannon entropy of completion against the length of said completion for (i) the ground truth, (ii) the base Llama model and (iii) the fine-tuned s1 model.}
    \label{fig:ent_vs_length}
\end{figure}

    \paragraph{Base vs fine-tuned models:} The base model produced visibly lower-quality completions compared to the fine-tuned models. In many cases, it appended (often erroneous) metadata or repeated text. Indeed, in our spot checks, we found several completions with repeated phrases. This prompted a more systematic analysis: we plotted the exponential of the Shannon entropy—computed from word frequency distributions as a function of text length—for the ground truth, the base model, and the fine-tuned models (see Figure~\ref{fig:ent_vs_length}). Lower entropy indicates repetition, while higher entropy indicates lexical variety.\footnote{For example, the entropy of “cat cat cat” is zero, as there is only one unique word, whereas that of “cat dog horse” is $\log(3)$ due to the presence of three unique words.} As expected, entropy increases with text length for the ground truth, reflecting increasing variability. In contrast, the base model often generates long completions with low entropy, indicating repetition. The fine-tuned models, however, exhibit far fewer such cases.

    Furthermore, with the fine-tuned models the text was logically coherent and the metadata at the end of the completions disappeared. We did not observe consistently better (in terms of human evaluations) completions in favor of either the LoRA-QKV or LoRA-all variants.

     \paragraph{Use of technical language:} The fine-tuned models could use technical language and, although training exclusively on abstracts limited their factual accuracy, demonstrated competitive performance relative to commercial LLMs. Table~\ref{tab:prompt-completions} demonstrates how two of our fine-tuned models (s3 and s8 LoRA-all) performed against the ground truth and ChatGPT-o. The fine-tuned models can clearly use technical language in sensible and relevant contexts, but are not 100\% factually correct. For instance, in this example, the s3 model sensibly refers to the class-$\mathcal S$ $N=2$ superconformal field theories, which arise in compactifications of M5-branes, while the s8 model similarly picks up on the related AGT correspondence. The completion of ChatGPT, on the other hand, is very close to the content of the actual paper referencing key components of the proof. The completion of the base Llama model repeats elements of the prompt, as well as a line of (false) metadata at the end.

     \paragraph{Comparison against commercial LLMs:} We observed  that our fine-tuned models were occasionally drawn towards the same points as popular commercial LLMs. For example, in the case of the famous ``TASI Lectures on D-branes" by  J.\ Polchinski hep-th/9611050, the abstract was accidentally split at the initial of one of the referenced authors, S.\ Chaudhuri. The resulting prompt and true completion appear in Table~\ref{tab:dbrane-example}. The majority of the fine-tuned models produced sensible outputs, but all notably started the completion by referencing a real string theorist whose first name correctly started with S and had produced related work. In addition, 4 of the LoRA-all models and 2 of the LoRA-QKV models referenced the name of Stephen Shenker; while this is not the correct completion, it was also the name referenced by ChatGPT-o, Claude-Sonnet 4, Gemini and DeepSeek.

\begin{table}[H]
\small
\centering
\begin{tabularx}{\textwidth}{>{\bfseries}p{2cm} X}
\toprule
 & {\bf Completions}  \\
\midrule
Prompt & It is known that the two- and three-point functions of Higgs-branch superconformal primaries in 4d $N=2$ superconformal field theories obey non-renormalization theorems on $N=2$ superconformal manifolds. In this paper we prove a stronger statement, that the bundles of Higgs-branch superconformal primaries over $N=2$ superconformal manifolds are endowed with a flat connection, or equivalently that Higgs-branch superconformal primaries have vanishing Berry phases under $N=2$ exactly marginal deformations. \\
\addlinespace
\midrule

True ~~~~~\text{completion} & This statement fits well with the proposed correspondence between the rigid structures of 2d chiral algebras and the sector of Schur operators in 4d $N=2$ theories. We also discuss the general interplay between non-renormalization theorems and the curvature of bundles of protected operators and provide a new simpler proof of the vanishing curvature of 1/2-BPS operators in 4d $N=4$ SYM theory that does not require the use of the 4d $tt^*$ equations.\\
\addlinespace
\midrule
\midrule
Llama & We also extend this result to the three-point functions of Higgs-branch superconformal primaries. The key ingredient in our proof is a non-renormalization theorem for the Higgs-branch superconformal primaries and the chiral ring relations. Our result has implications for the existence of $N=2$ superconformal manifolds in $N=2$ superconformal field theories. We also show that the Berry phases of the Coulomb-branch superconformal primaries do not vanish in general. From: Wenbin Yan [view email][v1] Fri, 28 Jun 2013 21:00:04 GMT (18kb) \\
\addlinespace
\midrule

s3 ~~~~~~LoRA-all & This is done by establishing a direct connection between the structure of the Higgs-branch chiral ring and the Higgs-branch superconformal primaries. The flat connection is shown to arise from the existence of an additional `dual' chiral ring, which is defined in terms of the Higgs-branch chiral ring. The flat connection is explicitly constructed for all $N=2$ superconformal field theories obtained from compactifications of M5-branes on Riemann surfaces. \\
\addlinespace
\midrule

s8 ~~~~~~LoRA-all &We also show that the Berry connection of a Higgs-branch primary is the same as the connection of the superconformal primary on the Coulomb branch, which is expected to be flat by the AGT correspondence. We give an explicit formula for the Berry connection in terms of the Coulomb branch superconformal primaries. \\
\addlinespace
\midrule

ChatGPT-o &We derive this result by analyzing the structure of the superconformal algebra and exploiting the shortening conditions obeyed by Higgs-branch operators. In particular, we show that the superconformal descendants required to generate a non-trivial Berry curvature are absent due to the protected nature of these multiplets. Furthermore, we demonstrate that the flatness of the associated bundles follows from supersymmetric Ward identities and constraints imposed by the $N=2$ superconformal symmetry. Our results imply that correlation functions of Higgs-branch primaries are covariantly constant across the conformal manifold, and we discuss the implications of this rigidity for the global geometry of the space of theories, including connections to $tt^*$ geometry and the absence of operator mixing. We conclude with comments on possible extensions to theories with additional supersymmetry and the role of Higgs-branch operators in the broader landscape of protected sectors in superconformal field theories. \\
\bottomrule
\end{tabularx}
\vspace{0.5em}
\caption{Prompt and abstract completions based on the paper 1807.04296, ``Geometry of Higgs-branch superconformal primary bundles'' by one of the authors, VN.}
\label{tab:prompt-completions}
\end{table}

     \paragraph{Creative completions:} Although fine-tuning our models exclusively on abstracts imposes clear limitations on their factual accuracy, we noticed several creative moments where they attempted interesting (and sometimes unexpected) connections to concepts outside the prompt that could be potentially relevant. For example, a prompt based on A.\ Sen's hep-th/9911116 paper on the universality of the tachyon potential in String Theory referred to the vanishing of the total energy in brane/anti-brane systems at the minimum of the tachyon potential. The s5 LoRA-all model used this fact to make an interesting connection with cosmology:  ``\ldots This suggests that the tachyon condensation in the background of a cosmological constant can result in its annihilation, providing a solution to the cosmological constant problem in a purely string theoretic context.'' Although this is a superficial statement, we found the attempt to make such a connection interesting.

     \paragraph{s10 models:} Our inspection of explicit samples confirmed that the s10 models, which were trained on the unorthodox combination of hep-th, q-bio and cs datasets, did not perform worse than other variants (in agreement with the picture arising from the analysis of the perplexities). 

In App.\ \ref{app:extra_completions} we further discuss some of these observations with a collection of a few additional explicit completions.

\begin{table}[t!]
\small
\centering
\begin{tabularx}{\textwidth}{>{\bfseries}p{3cm} X}
\toprule
\textbf{Component} & \textbf{Content} \\\midrule
Prompt & This is an introduction to the properties of D-branes, topological defects in string theory on which string endpoints can live. D-branes provide a simple description of various nonperturbative objects required by string duality, and give new insight into the quantum mechanics of black holes and the nature of spacetime at the shortest distances. The first two thirds of these lectures closely follow the earlier ITP lectures hep-th/9602052, written with S. \\
\addlinespace
\midrule

True completion & Chaudhuri and C. Johnson. The final third includes more extensive applications to string duality.\\

\bottomrule
\end{tabularx}
\vspace{0.5em}
\caption{Prompt and true completion for hep-th/9611050, ``TASI Lectures on D-Branes'' by J.\  Polchinski.}
\label{tab:dbrane-example}
\end{table}

\section{Discussion of results and future directions}

In this paper we presented results pertaining to the fine-tuning of  a collection of LLMs on the Llama 3.1 8B base model with the immediate goal of completing hep-th abstracts. All our models performed better than the base model, using appropriate technical language, despite the obvious shortcomings related to the fact that they have been trained only on abstracts and are based on very small models of only 8 billion parameters. We observed that augmenting the training set with other subject categories improved performance, even when these were not directly related to physics. This might be important for specialized LLMs in subject areas (like hep-th) with relatively small arXiv repositories. It would be interesting to investigate how this augmentation works when fine-tuning with whole papers, which would be the next logical step in this direction. 

There are many improvements that one could bring into the process we have reported on. Immediate developments could include more extensive hyperparameter tuning and training on carefully curated samples; see e.g. \cite{Chung:2025nsd} for applications in theoretical physics. Eventually, we would like to extend our models toward fully developed conversational agents, with an implementation of post-training Retrieval Augmented Generation (RAG) and reinforcement-learning steps to improve reasoning. Our ultimate goal is to develop a research assistant for high-energy theoretical physics. We hope to report related progress in the near future.

\section*{Acknowledgments}

We would like to thank Josh Nguyen and Yuan-Sen Ting for correspondence regarding the hyperparameters used to train AstroLlama. The work of CP was partially supported by the Science and Technology Facilities Council (STFC) Consolidated Grant ST/X00063X/1 “Amplitudes, Strings \& Duality”. The work of PR was partially supported by the Leverhulme Trust. This research utilised Queen Mary's Apocrita HPC facility, supported by QMUL Research-IT. \href{http://doi.org/10.5281/zenodo.438045}{http://doi.org/10.5281/zenodo.438045}. Calculations were also performed using the Sulis Tier 2 HPC platform hosted by the Scientific Computing Research Technology Platform at the University of Warwick. Sulis is funded by EPSRC Grant EP/T022108/1 and the HPC Midlands+ consortium. 

\section*{Disclaimer}
PA worked on this project in a personal capacity. The views and conclusions expressed are solely those of the authors and do not reflect those of Amazon. This work is independent of PA's professional duties and is neither affiliated with nor endorsed by Amazon. 

BC worked on this project in a personal capacity. The views and conclusions expressed are solely those of the authors and do not reflect those of Meta. This work is independent of BC's professional duties and is neither affiliated with nor endorsed by Meta.

\bibliography{LLM}

\providecommand{\href}[2]{#2}\begingroup\raggedright\begin{thebibliography}{10}

\bibitem{bommasani2022opportunitiesrisksfoundationmodels}
R.~Bommasani~{\it et al}, ``On the opportunities and risks of foundation
  models,'' \href{https://arxiv.org/abs/2108.07258}{{\ttfamily arXiv:2108.07258
  [cs.LG]}}.

\bibitem{openai2024gpt4technicalreport}
OpenAI and J.~Achiam~{\it et al}, ``{GPT-4 Technical Report},''
  \href{https://arxiv.org/abs/2303.08774}{{\ttfamily arXiv:2303.08774
  [cs.CL]}}.

\bibitem{touvron2023llamaopenefficientfoundation}
H.~Touvron~{\it et al}, ``{LLaMA: Open and Efficient Foundation Language
  Models},'' \href{https://arxiv.org/abs/2302.13971}{{\ttfamily
  arXiv:2302.13971 [cs.CL]}}.

\bibitem{touvron2023llama2openfoundation}
H.~Touvron~{\it et al}, ``{Llama 2: Open Foundation and Fine-Tuned Chat
  Models},'' \href{https://arxiv.org/abs/2307.09288}{{\ttfamily
  arXiv:2307.09288 [cs.CL]}}.

\bibitem{dubey2024llama3herdmodels}
A.~Dubey~{\it et al}, ``{The Llama 3 Herd of Models},''
  \href{https://arxiv.org/abs/2407.21783}{{\ttfamily arXiv:2407.21783
  [cs.AI]}}.

\bibitem{zhang2024comprehensivesurveyscientificlarge}
Y.~Zhang, X.~Chen, B.~Jin, S.~Wang, S.~Ji, W.~Wang, and J.~Han, ``A
  comprehensive survey of scientific large language models and their
  applications in scientific discovery,''
  \href{https://arxiv.org/abs/2406.10833}{{\ttfamily arXiv:2406.10833
  [cs.CL]}}.

\bibitem{bubeck2025earlysciaccelgpt5}
S.~Bubeck, C.~Coester, {\em et~al.}, ``Early science acceleration experiments
  with {GPT-5},'' \href{https://arxiv.org/abs/2511.16072}{{\ttfamily
  arXiv:2511.16072 [cs.CL]}}. \url{https://arxiv.org/abs/2511.16072}.

\bibitem{grezes2021buildingastrobertlanguagemodel}
F.~Grezes~{\it et al}., ``Building astrobert, a language model for astronomy
  and astrophysics,'' \href{https://arxiv.org/abs/2112.00590}{{\ttfamily
  arXiv:2112.00590 [cs.CL]}}.

\bibitem{Nguyen:2023nhp}
T.~D. Nguyen~{\it et al}, ``{AstroLLaMA: Towards Specialized Foundation Models
  in Astronomy},'' \href{https://arxiv.org/abs/2309.06126}{{\ttfamily
  arXiv:2309.06126 [astro-ph.IM]}}.

\bibitem{UniverseTBD:2024pmh}
{\bfseries {\rm UniverseTBD}} Collaboration, E.~Perkowski~{\it et al},
  ``{AstroLLaMA-Chat: Scaling AstroLLaMA with Conversational and Diverse
  Datasets},'' \href{https://dx.doi.org/10.3847/2515-5172/ad1abe}{{\em Res.
  Notes AAS} {\bfseries 8} no.~1, (2024) 7},
  \href{https://arxiv.org/abs/2401.01916}{{\ttfamily arXiv:2401.01916
  [astro-ph.IM]}}.

\bibitem{ting2024astromlab1winsastronomy}
Y.-S. Ting~{\it et al}, ``{AstroMLab 1: Who Wins Astronomy Jeopardy!?}''
  \href{https://arxiv.org/abs/2407.11194}{{\ttfamily arXiv:2407.11194
  [astro-ph.IM]}}.

\bibitem{perkowski2024astrollamachatscalingastrollamaconversational}
E.~Perkowski~{\it et al}, ``Astrollama-chat: Scaling astrollama with
  conversational and diverse datasets,''
  \href{https://arxiv.org/abs/2401.01916}{{\ttfamily arXiv:2401.01916
  [astro-ph.IM]}}.

\bibitem{dehaan2024cosmosagenaturallanguageassistantcosmologists}
T.~de~Haan, ``{cosmosage: A natural-language assistant for cosmology},''
  \href{https://dx.doi.org/10.1016/j.ascom.2025.100934}{{\em Astron. Comput.}
  {\bfseries 51} (2025) 100934},
  \href{https://arxiv.org/abs/2407.04420}{{\ttfamily arXiv:2407.04420
  [astro-ph.IM]}}.

\bibitem{hellert2024physberttextembeddingmodel}
T.~Hellert, J.~Montenegro, and A.~Pollastro, ``Physbert: A text embedding model
  for physics scientific literature,''
  \href{https://arxiv.org/abs/2408.09574}{{\ttfamily arXiv:2408.09574
  [physics.comp-ph]}}.

\bibitem{howard2023learningjumps}
J.~Howard, ``{Can LLMs learn from a single example?}'' Sept., 2023.
\newblock \url{https://www.fast.ai/posts/2023-09-04-learning-jumps/}.

\bibitem{lewkowycz2022solvingquantitativereasoningproblems}
A.~Lewkowycz~{\it et al}, ``Solving quantitative reasoning problems with
  language models,'' \href{https://arxiv.org/abs/2206.14858}{{\ttfamily
  arXiv:2206.14858}}.

\bibitem{dettmers2023case}
T.~Dettmers and L.~Zettlemoyer, ``The case for 4-bit precision: k-bit inference
  scaling laws,'' in {\em International Conference on Machine Learning},
  pp.~7750--7774, PMLR.
\newblock 2023.

\bibitem{dettmers2023qlora}
T.~Dettmers, A.~Pagnoni, A.~Holtzman, and L.~Zettlemoyer, ``{QLoRa: Efficient
  Finetuning of Quantized LLMs},''
  \href{https://arxiv.org/abs/2305.14314}{{\ttfamily arXiv:2305.14314
  [cs.LG]}}.

\bibitem{hu2022lora}
E.~Hu~{\it et al}, ``{Lo{RA}: Low-Rank Adaptation of Large Language Models},''
  in {\em International Conference on Learning Representations}.
\newblock 2022.
\newblock \url{https://openreview.net/forum?id=nZeVKeeFYf9}.

\bibitem{dao2023flashattention2}
T.~Dao, ``Flash{A}ttention-2: Faster attention with better parallelism and work
  partitioning,'' in {\em International Conference on Learning Representations
  (ICLR)}.
\newblock 2024.

\bibitem{huyen2024ai}
C.~Huyen, {\em {AI Engineering: Building Applications with Foundation Models}}.
\newblock O’Reilly Media, Sebastopol, CA, Dec., 2024.

\bibitem{dai2019transformerxl}
Z.~Dai, Z.~Yang, Y.~Yang, J.~Carbonell, Q.~V. Le, and R.~Salakhutdinov,
  ``Transformer-xl: Attentive language models beyond a fixed-length context.''
  2019.
\newblock \url{https://arxiv.org/abs/1901.02860}.

\bibitem{2024arXiv240117072A}
A.~{Aynetdinov} and A.~{Akbik}, ``{SemScore: Automated Evaluation of
  Instruction-Tuned LLMs based on Semantic Textual Similarity},''
  \href{https://arxiv.org/abs/2401.17072}{{\ttfamily 2401.17072 [cs.CL]}}.

\bibitem{koo2016guideline}
T.~K. Koo and M.~Y. Li, ``A guideline of selecting and reporting intraclass
  correlation coefficients for reliability research,'' {\em Journal of
  Chiropractic Medicine} {\bfseries 15} no.~2, (2016) 155--163.

\bibitem{Chung:2025nsd}
D.~J.~H. Chung, Z.~Gao, Y.~Kvasiuk, T.~Li, M.~Münchmeyer, M.~Rudolph, F.~Sala,
  and S.~C. Tadepalli, ``Theoretical physics benchmark (tpbench)—a dataset
  and study of ai reasoning capabilities in theoretical physics,''
  \href{https://dx.doi.org/10.1088/2632-2153/adfcb0}{{\em Machine Learning:
  Science and Technology} {\bfseries 6} no.~3, (Sep, 2025) 030505}.
  \url{https://dx.doi.org/10.1088/2632-2153/adfcb0}.

\bibitem{mann1947}
H.~B. Mann and D.~R. Whitney, ``On a test of whether one of two random
  variables is stochastically larger than the other,'' {\em Ann. Math.
  Statist.} {\bfseries 18} (1947) 50--60.

\bibitem{kerby2014}
D.~S. Kerby, ``The simple difference formula: An approach to teaching
  nonparametric correlation,'' {\em Compr. Psychol.} {\bfseries 3} (2014)
  11.IT.3.1.

\bibitem{cohen1988}
J.~Cohen, {\em Statistical Power Analysis for the Behavioral Sciences}.
\newblock Lawrence Erlbaum Associates, Hillsdale, NJ, 2nd~ed., 1988.

\bibitem{spearman1904}
C.~Spearman, ``The proof and measurement of association between two things,''
  {\em Am. J. Psychol.} {\bfseries 15} (1904) 72--101.

\bibitem{shrout1979}
P.~E. Shrout and J.~L. Fleiss, ``Intraclass correlations: Uses in assessing
  rater reliability,'' {\em Psychol. Bull.} {\bfseries 86} (1979) 420--428.

\bibitem{koo2016}
T.~K. Koo and M.~Y. Li, ``A guideline of selecting and reporting intraclass
  correlation coefficients for reliability research,'' {\em J. Chiropr. Med.}
  {\bfseries 15} (2016) 155--163.

\end{thebibliography}\endgroup
\bibliographystyle{utphys}

\vskip 1cm
\appendix

\addtocontents{toc}{\protect\setcounter{tocdepth}{0}}

{\huge Appendix}

\section{Data Curation}\label{curation}

Following extraction, the data was cleaned by removing records where the word `withdrawn' appeared in either the `comments' or `abstract' data field. All remaining entries in `abstract' had linebreaks and following whitespaces replaced with a single whitespace character. The abstracts also had any leading/trailing whitespace stripped.

For the tokenization of the data we first set the tokenizer's padding token to be the end of sequence (EOS) token. We then tokenized the `abstract' column of the train and test datasets without padding or truncation. To automatically handle the batching of the tokenized datasets we employed the DataCollatorForSeq2Seq data collator class from the transformers library. This data collator pads each of the 16 sequences in a batch to be of the same length as the longest sequence. It also correctly applies an attention mask to each padded sequence, attending to the first padding token, which represents the EOS token, and ignoring the remaining ones. As part of these evaluations, we also used the control dataset of five abstracts as a qualitative guide to training improvement. 

The precise composition of the datasets used for our fine-tuning follows: 

{\bf s1}: Consists of papers whose primary arXiv listing is hep-th. 

{\bf s2}: Consists of papers whose primary arXiv listing is either hep-ph or gr-qc with ratios: hep-ph 67.58\% and gr-qc 32.42\%.

{\bf s3}: Consists of the concatenation of s1 and s2 followed by random reshuffle.

{\bf s4}: Consists of a 70\% hep-th sample, and then 15\% each of the hep-ph and gr-qc entries in the s3 dataset, all randomly shuffled.

{\bf s5}: Consists of a 85\% hep-th sample and 15\% of gr-qc entries in the s3 dataset, concatenated and randomly shuffled.

{\bf s6}: Consists of a 85\% hep-th sample and 15\% of hep-ph entries in the s3 dataset, concatenated and randomly shuffled.

{\bf s7}: Consists of the concatenation of the  hep-th and gr-qc datasets (not from s3) followed by random reshuffle.

{\bf s8}: Consists of the concatenation of the hep-th and hep-ph datasets (not from s3) followed by random reshuffle.

{\bf s9}: Consists of a combination of gr-qc and hep-ph (not from s3) but is limited to having same size as hep-th.
We take a proportion given by
\begin{equation*}
    \frac{\text{size of hep-th}}{\text{size of hep-ph} + \text{size of gr-qc}} = 0.5379232194539302
\end{equation*}
from each of the hep-ph and gr-qc datasets. We then concatenate and reshuffle.

{\bf s10}: Consists of the concatenation of hep-th, q-bio datasets with padding from cs to make dataset same size as s3. \\

\section{Training Hyperparameters and Inference Settings}\label{fullhypers}

For full reproducibility, we provide complete specifications of all training hyperparameters and inference settings. These expand upon the summary provided in the main text, Table \ref{tab:training-config}. Table~\ref{tab:training_hyperparameters} presents the complete training configuration used for fine-tuning the Llama-3.1-8B base model with LoRA. Table~\ref{tab:generation_params} presents the complete decoding configuration used during inference. Parameters not explicitly configured used HuggingFace Transformers library defaults (v4.45.2).

\begin{table}[htbp]
\centering
\renewcommand{\arraystretch}{1.3}
\begin{tabular}{@{}ll@{}}
\toprule
\textbf{Parameter} & \textbf{Value} \\
\midrule
\multicolumn{2}{l}{\textit{Model Configuration}} \\
Base model & meta-llama/Meta-Llama-3.1-8B \\
Attention mechanism & Flash Attention 2 \\
Quantization & 4-bit NF4 with double quantization \\
Compute dtype & bfloat16 \\
\midrule
\multicolumn{2}{l}{\textit{LoRA Configuration}} \\
Rank (r) & 8 \\
Alpha & 32 \\
Dropout & 0.05 \\
Target modules & q, k, v, o, up, down, gate projections \\
\midrule
\multicolumn{2}{l}{\textit{Training Configuration}} \\
Training epochs & 4 \\
Batch size (per device) & 16 \\
Gradient accumulation & 1 \\
Max gradient norm & 1.0 \\
Mixed precision & bfloat16 \\
\midrule
\multicolumn{2}{l}{\textit{Optimizer (AdamW)}} \\
Learning rate & 3e-4 \\
Betas & [0.9, 0.95] \\
Epsilon & 1e-5 \\
Weight decay & 0.1 \\
\midrule
\multicolumn{2}{l}{\textit{Learning Rate Scheduler}} \\
Type & Cosine with minimum LR \\
Warmup ratio & 0.1 \\
Minimum LR rate & 0.01 (1\% of initial LR) \\
Minimum LR (absolute) & 3e-6 \\
\midrule
\multicolumn{2}{l}{\textit{Dataset}} \\
Training data & LLMsForHepth/hep-th\_primary (train split) \\
Tokenizer & meta-llama/Meta-Llama-3.1-8B \\
Random seed & 42 \\
\bottomrule
\end{tabular}
\vskip 0.25cm
\caption{Training Hyperparameters}
\label{tab:training_hyperparameters}
\renewcommand{\arraystretch}{1.0}
\end{table}

\begin{table}[htbp]
\centering
\renewcommand{\arraystretch}{1.3}
\begin{tabular}{@{}lll@{}}
\toprule
\textbf{Parameter} & \textbf{Value} & \textbf{Source} \\
\midrule
\multicolumn{3}{l}{\textit{Model Configuration}} \\
Attention mechanism & SDPA & Configured \\
Compute dtype & float16 & Configured \\
\midrule
\multicolumn{3}{l}{\textit{Generation Configuration}} \\
max\_new\_tokens & 1024 & Configured \\
min\_new\_tokens & 1 & Configured \\
temperature & 0.7 & Configured \\
do\_sample & true & Configured \\
top\_p & 0.9 & Default \\
top\_k & 50 & Default \\
repetition\_penalty & 1.0 & Default \\
no\_repeat\_ngram\_size & 0 & Default \\
Stop criteria & EOS token & Default \\
Random seed & 42 & Configured \\
Padding side & left & Configured \\
Cache implementation & static & Configured \\
\bottomrule
\end{tabular}
\vskip 0.25cm
\caption{Text Generation Parameters}
\label{tab:generation_params}
\renewcommand{\arraystretch}{1.0}
\end{table}

\section{Cross Entropy Loss and Perplexity}\label{app:cel_perplexity}

Causal language models predict the probabilities of the next word following a sequence of words from those in the dictionary
\be
p_\theta(w_{t+1}| w_{1:t})\;,
\ee
where $\theta$ are the parameters of the model and $w_k \in \mathcal V$, where $\mathcal V$ is the whole vocabulary.  

During pre-training and fine-tuning, we have access to the ground truth words at every location $y_k$. The training is done by maximizing the likelihood of the ground truth data by minimizing the cross entropy loss. The loss for a given sequence of $T$ words is given by
\be
\mathcal L = -\frac{1}{T} \sum_{t=0}^T \sum_{i \in |\mathcal V|} y_{t+1,i} \log(p_\theta(w_{t+1,i}| w_{1:t}))\;.
\ee 
Of course, training happens not sequence by sequence but in batches. A given batch may end in the middle of a sequence, so the next batch may start mid-sequence as well. Since the loss is just an arithmetic mean over all token losses, we can concatenate sequences to improve efficiency as long as the language model uses correct context for each prediction. This requires proper causal masks that restrict attention to the causal past within the same sequence.

In practice, maintaining context across sequences that are split across elements of a batch is hard to implement. At the same time, doing so is more important in fine-tuning due to sparsity of data. Thus, in this paper we have opted to have each batch element have only one sequence and used padding to mask out the positions in the batch elements after the sequence.

While training is done by minimizing cross entropy loss, a common metric in Natural Language Processing is perplexity. For a given sequence, it is simply the exponential of the cross entropy loss
\be
\mathcal P = \Bigg( \prod_{t=0}^T  \frac{1}{p_\theta(w_{t+1,\text{Id}(y_{t+1})}| w_{1:t})} \Bigg)^{1/T}\;,
\ee
where $\text{Id}(y_k)$ denotes the index of the k-th ground truth word in the full vocabulary.

To understand the meaning of this metric, let us consider a toy language model that overfits on the training dataset. First consider the training dataset composed of one sentence "I love cats". Then if we start off with the first word "I" as a prompt, the language model will be sure the next word is "love" so the probability will be peaked on that word at 1 and 0 for all others. Then given the partial sentence "I love" the next word will similarly be peaked at "cats" with probability 1 and 0 for all others. The perplexity will be 1. Now consider adding a sentence "I love dogs" to the training set. Suppose we are evaluating against "I love cats". Now we again start off with "I" and the language model will be sure of the next word being "love" but then it will have $\frac{1}{2}$ probabilities for "cats". The unnormalized perplexity will be 2. Now add another sentence "I adore Llamas". We continue evaluating against "I love cats". Now when we start off with "I", the model will predict "love" as the next word with probability $\frac{2}{3}$. Once it has that, it will predict "cats" with probability $\frac{1}{2}$. The unnormalized perplexity will be 3. In general, the unnormalized perplexity counts the possible number of ways a sequence branches off. For concreteness, let us compute the perplexities of the three sequences in our toy dataset:

\begin{itemize}
\item "I love cats": (X,2/3,1/2)  $\rightarrow 3$ 
\item "I love dogs": (X,2/3,1/2) $\rightarrow 3$
\item "I adore Llamas" : (X,1/3,1) $\rightarrow 3$
\end{itemize}

The fact that all three ended up with the same perplexity of 3 is purely stochastic. Let us see what happens when we do a sentence completion with the context length set to two words:

\begin{itemize}
\item "I love cats": (X,X,1/2)  $\rightarrow 2$
\item "I love dogs": (X,X,1/2) $\rightarrow 2$
\item "I adore Llamas" : (X,X,1) $\rightarrow 1$
\end{itemize}

What we just discussed is the ground truth perplexity coming from the fact that the training data itself has ambiguity in how a sequence is completed having started off with "I". This is the lowest perplexity achievable. Models in general will have higher perplexity. The normalization is done so as able to compare perplexities across sequences of different lengths.

The above discussion involved the perplexity evaluated against one ground truth sequence. How do we compute it for a set of sequences? A naive answer, commonly used in the literature as it is computationally simpler, would be to take the cross entropy loss for the whole dataset and exponentiate it. This would amount to a geometric mean of perplexities of sequences. Another possibility is to take the arithmetic mean of perplexities of the sequences as this would give the average number of branching factors given a prompt. 

In the toy example above, for the case of context length being a single word, both the arithmetic and geometric mean are $3^{1/3}$. For the case of the context length being two words, the geometric mean is $2^{2/9}$ while the arithmetic mean is $\frac{2 \times 2^{1/3} + 1}{3}$. While both metrics have theoretical merit, most implementations default to the geometric mean approach since it naturally emerges from standard training procedures where cross-entropy losses are averaged before exponentiation. However, the arithmetic mean may be more interpretable when comparing model performance across datasets with varying sequence lengths, as it directly reflects the average branching factor.

\section{Additional examples of completions}
\label{app:extra_completions}

In this appendix, we collect a couple of additional completions to demonstrate more explicitly some of the observations that were summarized in Section \ref{human}.

In Section \ref{human}, we pointed out that the overall quality of the completions of the base model was worse compared to that of the fine-tuned models and the plots of Figure\ \ref{fig:ent_vs_length} re-affirmed that the base model exhibited longer completions with repeated text.

For example, when asked to complete a prompt based on the well-known paper hep-th/9503124, `String Theory Dynamics In Various Dimensions' by E.\ Witten, the base model performed in the following manner:

\noindent
{\textbf{Prompt}:} The strong coupling dynamics of string theories in dimension $d\geq 4$ are studied.

\noindent
{\textbf{True completion}:} It is argued, among other things, that eleven-dimensional supergravity arises as a low energy limit of the ten-dimensional Type IIA superstring, and that a recently conjectured duality between the heterotic string and Type IIA superstrings controls the strong coupling dynamics of the heterotic string in five, six, and seven dimensions and implies $S$ duality for both heterotic and Type II strings.

\noindent
{\textbf{Base-model completion}:} We show that, for any theory with a weakly curved target space, the effective action is well approximated by a local, conformally invariant, derivative expansion. In particular, the coupling of matter fields to the metric, the connection and the dilaton is described by a universal effective action of a given form, with coefficients that depend only on the central charges of the theory. We also show that the couplings to higher form gauge fields are governed by a similar universal effective action. These results are used to study the gravitational and gauge field couplings of a class of four dimensional string theories. We show that the gravitational couplings of all theories are governed by the same action. We also show that, for a wide class of theories, the gauge field couplings are governed by the same effective action. The only difference between the theories is in the gauge field kinetic terms and in the structure of the anomaly terms. We also study the effective action of a class of higher dimensional string theories and show that, for a wide class of theories, the effective action is governed by the same general form, with coefficients that depend only on the central charges of the theory. We show that the gravitational and gauge field couplings of these theories are described by the same effective action, up to the same structure of anomaly terms. The only difference between the theories is in the gauge field kinetic terms.
From: Ashoke Sen [ view email ]
[v1] Tue, 18 Mar 2003 16:04:08 GMT (24kb)

The completions of the fine-tuned models, on the other hand, are more concise, much less repetitive and do not exhibit false metadata at the end. For example, the completion of the s5 LoRA-all model is:

\noindent
{\textbf{s5 LoRA-all completion}:} It is shown that for $d\geq 4$ the effective action is invariant under the T-duality group $O(d,d;\mathbb{Z})$. The T-duality group $O(d,d;\mathbb{Z})$ acts on the moduli space of the effective action in the same way as the strong-weak coupling duality group $SL(2,\mathbb{Z})$ acts on the moduli space of the effective action in dimension $d=2$.

In the main text, we also pointed out that the s10 models, which were trained on the unorthodox combination of hep-th, q-bio and cs datasets do not, in general, perform worse than the other variants. Indeed, in this example, the completion of the s10 LoRA-all model is also short without repetitions, and includes an interesting hint to quantum gravity and the structure of the string spectrum:

\noindent
{\textbf{s10 LoRA-all completion}:} We argue that the string coupling constant becomes effectively weak when the distance from the string scale exceeds the distance between the lightest string excitations. This effect is a consequence of the fact that the spectrum of string excitations in $d\geq 4$ is discrete and has a finite number of states in any finite volume. We discuss the relevance of this effect to the problem of quantum gravity.

\section{Evaluation Metrics}\label{statmetrics}

In this appendix we provide a qualitative description of the metrics used to assess the human evaluation part of this manuscript. 

\subsection{Statistical Comparison Metrics}
\begin{itemize}
    \item \textbf{Mann-Whitney U Test}~\cite{mann1947}: A non-parametric test that compares scores between two groups (e.g., fine-tuned vs.\ commercial models) to determine if one systematically scores higher than the other, without assuming that scores follow a Gaussian distribution. The test works by pooling all scores from both groups, ranking them, and computing the \textbf{U statistic}---essentially the number of times a score from group A exceeds a score from group B. For intuition: if group A consistently outperforms group B, most A scores will rank above most B scores, yielding a small $U$ (or equivalently, a large $U$ for the reverse comparison). The $U$ statistic alone is difficult to interpret directly; instead, one examines the $p$-value and effect size.
    \item \textbf{p-value}: The probability of seeing a difference this large by chance alone. Values below 0.05 are conventionally considered statistically significant.
    \item \textbf{Effect Size (r)}: The rank-biserial correlation~\cite{kerby2014}, measuring how large the difference is between groups on a scale from 0 to 1. This is a useful complement to the $p$-value, since large samples can yield small $p$-values even for negligible real-world differences. Interpretation: $|r| < 0.3$ (small), $0.3 \leq |r| < 0.5$ (medium), $|r| \geq 0.5$ (large)~\cite{cohen1988}.
\end{itemize}

\subsection{Inter-Rater Reliability Metrics}
\begin{itemize}
    \item \textbf{Spearman $\rho$}~\cite{spearman1904}: Measures whether two raters rank completions similarly (e.g., both rank completion A above B), even if they use different parts of the 1--10 scale. Values near 1 indicate strong agreement on relative ordering.
    \item \textbf{ICC (Intraclass Correlation)}~\cite{shrout1979}: Measures how consistently raters assign the same absolute scores, not just the same rankings. ICC(2,1) quantifies single-rater reliability; ICC(2,k) quantifies the reliability of the average across $k$ raters. Values above 0.75 are generally considered good, and above 0.5 moderate~\cite{koo2016}.
    \item \textbf{Agreement Rates}: Measures the fraction of cases where raters gave the same score (exact), or scores differing by at most $\pm 1$ or $\pm 2$ points.
\end{itemize}

\end{document}